# Domain Knowledge Uncertainty and Probabilistic Parameter Constraints

**Yi Mao**
College of Computing
Georgia Institute of Technology
yi.mao@cc.gatech.edu

**Guy Lebanon**
College of Computing
Georgia Institute of Technology
lebanon@cc.gatech.edu

## Abstract

Incorporating domain knowledge into the modeling process is an effective way to improve learning accuracy. However, as it is provided by humans, domain knowledge can only be specified with some degree of uncertainty. We propose to explicitly model such uncertainty through probabilistic constraints over the parameter space. In contrast to hard parameter constraints, our approach is effective also when the domain knowledge is inaccurate and generally results in superior modeling accuracy. We focus on generative and conditional modeling where the parameters are assigned a Dirichlet or Gaussian prior and demonstrate the framework with experiments on both synthetic and real-world data.

## 1 Introduction

Incorporating domain knowledge into the modeling process is an effective way to improve learning accuracy. In some cases the knowledge may be incorporated by modifying the underlying statistical model. In most cases, however, standard off-the-shelf models are used such as logistic regression, SVM, mixture of Gaussians, etc., and the domain knowledge is integrated into the training process of these models by constraining the parameters to a certain region.

For example in document analysis, vocabulary words are treated as abstract orthogonal dimensions. The statistical relationship among the words and between the words and the predictor variable is determined solely based on the available data. Modifying the learning process so that it takes into consideration domain knowledge can substantially improve accuracy, especially when the training data is scarce e.g., [8, 9].

A fundamental difficulty with incorporating domain knowledge is that as it is provided by humans, it often holds with some degree of uncertainty. For example in sentiment prediction, the presence of the word `good` corresponds usually, but not always to a positive opinion. While this difficulty applies to explicitly formulated domain knowledge, it is even more pronounced when the domain knowledge is obtained implicitly by interpreting user feedback. For example in web search, clickthrough data or the time a user spent in a site are usually interpreted as indicating high relevance. This interpretation is correct in many but not all cases.

In this paper, we propose to explicitly model domain knowledge uncertainty by specifying the probability with which it is expected to hold. Specifically, we consider the case of a hierarchical prior over the parameter space with additional parameter constraints holding with certain probabilities. Thus in the case of $x \sim p(\cdot|\theta)$, $\theta \sim p(\cdot|\alpha)$ we enforce probabilistic parameter constraint $P(\theta \in A) \geq \eta$ where $A$ is a set corresponding to the domain knowledge and $\eta$ corresponds to the uncertainty or confidence level. We derive an equivalence between the probabilistic constraint $P(\theta \in A) \geq \eta$ and certain hard constraint over the hyperparameters $\alpha$. Inference can then proceed on the equivalent model using standard techniques such as empirical Bayes or maximum posterior estimate.

Our proposed framework applies to a large class of practical models. We focus on generative and conditional modeling where the parameters are assigned a Dirichlet or Gaussian prior. This includes the popular cases of ridge regression, mixture of Gaussians, regularized logistic regression, naive Bayes and smoothed $n$-gram estimation. We show that in these cases the framework translates into well defined and computable hyperparameter constraints and discuss computational schemes for performing Bayesian inference.

From a Bayesian perspective, our framework derives a prior consistent with uncertain domain knowledge and thus may be considered a form of prior elicitation. Its practical significance is that it enables the use of a large quantity of somewhat inaccurate knowledge which is otherwise problematic to use.

## 2 Probabilistic Constraints in Hierarchical Bayes

We consider situations in which the model is a hierarchical Bayes model

$$\begin{aligned} z &\sim f(\cdot|\theta) \qquad \theta \in \mathbb{R}^n \\ \theta &\sim g(\cdot|\alpha) \\ \alpha &\sim h(\cdot) \end{aligned} \tag{1}$$

where $f, g$ are distributions parameterized by $\theta, \alpha$ and $h$ is a hyperprior for $\alpha$. Abusing notation slightly, we consider the distribution $f$ to be over $z = x$ in the generative case i.e., $f(x|\theta)$, or a conditional model over $z = y|x$ in a discriminative setting i.e., $f(y|x, \theta)$. Model (1) is fairly standard and contains a wide variety of popular generative and conditional models such as regularized logistic regression, ridge regression and lasso, mixture of Gaussians, etc. In some cases the distribution $h(\alpha)$ is uniform or an uninformative prior. In other cases it is replaced with a fixed value altogether.

We introduce domain knowledge into the model by identifying sets $A_i, i = 1, \ldots, l$ which are expected to contain the parameters $\theta \in A_i$ with some degree of confidence. A simple case is linear constraints

$$A_i = \{\theta : a_i^\top \theta \leq b_i\} \quad a_i \in \mathbb{R}^n, b_i \in \mathbb{R} \tag{2}$$

which despite its simplicity is general enough to account for many practical situations. Some useful special cases that are achievable using (2) are

$$\theta_{\pi(1)} \leq \cdots \leq \theta_{\pi(k)} \tag{3}$$
$$b \leq \theta_i \leq c \tag{4}$$
$$b \leq |\theta_i - \theta_j| \leq c \tag{5}$$
$$b \leq \sum \theta_i \leq c. \tag{6}$$

Equation (3) represents a case where we know some parameters are likely to be larger than others ($\pi$ is a permutation over $n$ letters and $k < n$). Equation (4) represents a case where we know the parameter values are bounded, for example in logistic regression we might know that some parameters are positive $\theta_i \geq 0$ (contributing to positive class label) and some are negative $\theta_i \leq 0$ (contributing to negative class label). Equation (5) represents knowledge that two parameters are similar in value and Equation (6) determines that the total parameter value is somehow bounded.

The constraints $\theta \in A_i$ are assigned confidence values $\eta_i$ and incorporated into the model by pairing (1) with

$$\int_{A_i} g(\theta|\alpha)\, d\theta \geq \eta_i \quad i = 1, \ldots, l. \tag{7}$$

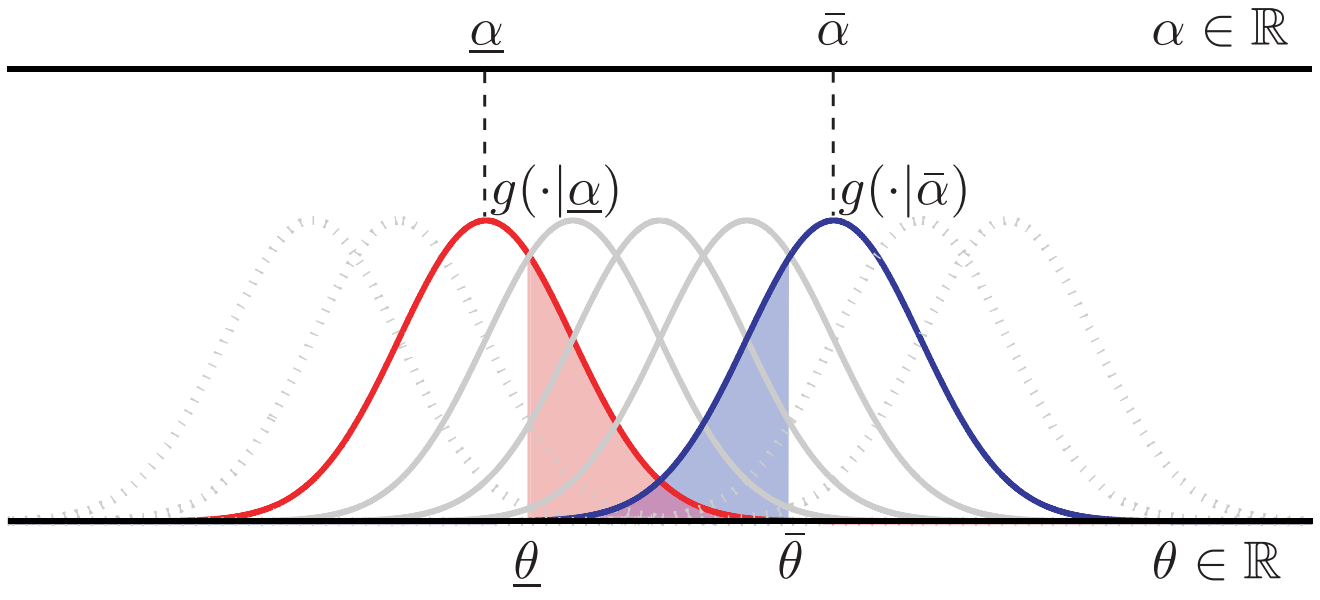

Figure 1: Illustration of proof for Proposition 1. $A_i$ is chosen to be $[\underline{\theta}, \bar{\theta}]$. For $\alpha \in [\underline{\alpha}, \bar{\alpha}]$, $\int_{A_i} g(\theta|\alpha)\, d\theta \geq \eta_i$, which implies $B_i = [\underline{\alpha}, \bar{\alpha}]$. Solid lines represent $g(\cdot|\alpha)$ for $\alpha \in B_i$ while dashed lines represent $\alpha \notin B_i$.

It is important to note that the constraints (7) may or may not be satisfied depending on the value of $\alpha$. If $\alpha$ is a fixed parameter the constrained problem is trivial - either (7) is satisfied or not. In the former case we can proceed with normal Bayesian inference and in the latter case we need to modify either the constraints (7) or the model (1). However, the situation gets more interesting when $\alpha$ is a random variable. In this case, standard Bayesian inference is modified to account for the constraints, effectively introducing the domain knowledge into the modeling process.

**Proposition 1.** *The model* (1) *subject to the constraints* (7) *is equivalent to the following Bayes model*

$$\begin{aligned} z &\sim f(\cdot|\theta) \\ \theta &\sim g(\cdot|\alpha) \\ \alpha &\sim c\, h(\cdot)\, 1_{\{\alpha \in B_1 \cap \cdots \cap B_l\}} \end{aligned} \tag{8}$$

*where $c$ ensures normalization and*

$$B_i = \left\{ \alpha : \int_{A_i} g(\theta|\alpha)\, d\theta \geq \eta_i \right\}.$$

*Proof.* The equivalence follows from considering separately the cases when the constraints are satisfied and when they are not (see Figure 1). □

The equivalence derived in Proposition 1 is useful as (8) is an unconstrained Bayesian model on which inference can proceed as usual, assuming the sets $B_1, \ldots, B_l$ are determined and $\cap_i B_i \neq \emptyset$. Specifically, assuming a dataset $\mathcal{D} = \{z^{(1)}, \ldots, z^{(m)}\}$, the full Bayesian treatment suggests integrating over the posterior to obtain expectations of interest. We focus on two alternatives due to computational consideration: empirical Bayes and maximum posterior.

In the case of empirical Bayes (EB) we obtain a point estimate for $\alpha$ by maximizing the posterior $p(\alpha|\mathcal{D})$

$$\alpha^* = \arg\max_{\alpha} \; h(\alpha) \int f(\mathcal{D}|\theta)g(\theta|\alpha)\,d\theta \quad \text{subject to } \alpha \in B_1 \cap \cdots \cap B_l \tag{9}$$

and use $\alpha^*$ to compute probabilities of interest. For example, we can classify a new example $x$ by

$$\hat{y} = \arg\max_{y} \int_{\theta} f(y|x,\theta)g(\theta|\alpha^*)\,d\theta.$$

A second alternative that may be used when the integration (9) is computationally intractable is maximum posterior (MAP) where $p(\alpha,\theta|\mathcal{D})$ is maximized to obtain point estimates for both $\alpha,\theta$

$$(\alpha^*,\theta^*) = \arg\max_{\alpha,\theta} \; f(\mathcal{D}|\theta)g(\theta|\alpha)h(\alpha) \quad \text{subject to } \alpha \in B_1 \cap \cdots \cap B_l. \tag{10}$$

In this case new examples may be classified as

$$\hat{y} = \arg\max_{y} f(y|x,\theta^*).$$

In general, it is often hard to invert the constraints (7) and obtain the sets $B_1,\ldots,B_l$ in Proposition 1. In the next two sections we derive the inversion for the case of linear constraints with either a Dirichlet or a Gaussian prior. The maximization problems (9), (10) may be solved using standard interior point optimization.

## 3 Dirichlet Prior

Dirichlet prior $g(\theta|\alpha)$ applies to a variety of models $f(z|\theta)$ whose parameters take values in the simplex

$$\theta \in \mathbb{P}_{n-1} = \left\{\theta \in \mathbb{R}^n : \theta_i \geq 0, \sum \theta_i = 1\right\}. \tag{11}$$

In particular, it is often used in conjunction with a multinomial $f(z|\theta)$ modeling the appearance of words or short phrases called $n$-grams. The MAP estimate for $f(z|\theta) = \text{Mult}(\theta)$, $g(\theta|\alpha) = \text{Dir}(\alpha)$ modifies the observed word counts by adding $\alpha_i$ to the count of word $i$ in the text and re-normalizing the modified count vector to form a probability distribution. Such models serve a key role in a wide variety of text processing tasks including language modeling, topic analysis, text classification, and syntactic parsing.

Since each dimension in the parameter vector $\theta$ corresponds directly to the probability that a certain word or phrase appears, it is easy to construct constraints $\theta \in A_i$ that correspond to linguistic knowledge. In the generative case, such knowledge may correspond to the identification of words that are more popular than others. For example, the following constraint may correspond to plausible linguistic knowledge

$$\theta_i \geq \theta_j \text{ if word } i \text{ is much shorter than word } j. \tag{12}$$

Such a statement may often hold as very long words tend to be uncommon and very short words tend to be common. However, as (12) is not always true it is best to enforce it with some confidence $\eta_i < 1$ in order to prevent poor estimation quality.

As a second example consider the conditional case where different multinomial models with Dirichlet priors are built separately for different class labels $y$. In this case domain knowledge may reflect the relationship between the class label and the words, in addition to the relationship among the words as in (12). For example, consider the case where the label $y$ corresponds to spam or not spam. It is relatively easy to come up with a list of keywords affiliated with spam emails (`free, information, $`) and constrain the corresponding $\theta_i$ to be large if the label $y$ equals spam and small otherwise. Such domain knowledge, while plausible, may not hold always and enforcing it categorically may result in poor estimation quality. On the other hand, enforcing the constraints with confidence $\eta_i < 1$ will allow the model to use the constraints when they apply and avoid them when they do not.

As mentioned in the previous section we focus in this paper on linear constraints (2). Such constraints are relatively flexible and they are able to capture ordered and axis aligned constraints (3)-(4) which include the two examples presented above as well as additional special cases such as (5)-(6).

The key to inverting the linear constraints (7) and identifying the sets $B_i$ in the case of a Dirichlet prior is the observation that if $X_j \sim \chi^2_{d_j}, j = 1,\ldots,n$ ($\chi^2_{d_j}$ represent independent chi-squared variables with $d_j$ degrees of freedom) then

$$\left(\frac{X_1}{\sum X_i},\ldots,\frac{X_n}{\sum X_i}\right) \sim \text{Dir}\left(\frac{d_1}{2},\ldots,\frac{d_n}{2}\right).$$

It follows that if $\theta \sim \text{Dir}(\alpha_1,\ldots,\alpha_n)$, we may construct independent random variables $Y_j \sim \chi^2_{2\alpha_j}$ so that

$$P\left(\sum_{j=1}^n a_j\theta_j \leq b\right) = P\left(\frac{\sum_j a_j Y_j}{\sum_j Y_j} \leq b\right) = P\left(\sum_j (a_j - b)Y_j \leq 0\right). \tag{13}$$

If $\lambda_1,\ldots,\lambda_u$ are $u$ distinct non-zero values of $a_j - b$, $j = 1,\ldots,n$, and $T_k \sim \chi^2_{r_k}$ with $r_k \stackrel{\text{def}}{=} 2\sum_j \alpha_j \delta(a_j - b, \lambda_k)$,

$k=1,\dots,u$, (13) becomes equivalent to

$$P\left(\sum_{k=1}^{u}\lambda_k T_k \le 0\right) = \frac{1}{2} - \frac{1}{\pi}\int_0^\infty \frac{\sin\left(\frac{1}{2}\sum_{k=1}^{u} r_k \tan^{-1}(\lambda_k t)\right)}{t\prod_{k=1}^{u}\left(1+\lambda_k^2 t^2\right)^{r_k/4}}dt \quad (14)$$

which is a function of $r_1,\dots,r_u$ and thus of $\alpha$[10].

Solving (14) is a difficult problem since it involves integration over a complex expression of $r_k$ which in turn depend on $\alpha$. We suggest to use the Edgeworth expansion to approximate (14). The Edgeworth expansion states that if $X$ is a random variable with finite moments, mean zero and variance one, then its density function $f$ can be approximated as either (15) or (16)

$$\frac{f(x)}{\phi(x)} \approx 1 + H_3(x)\frac{\kappa_3}{6} \quad (15)$$

$$\frac{f(x)}{\phi(x)} \approx 1 + H_3(x)\frac{\kappa_3}{6} + H_4(x)\frac{\kappa_4}{24} + H_6(x)\frac{\kappa_3^2}{72}. \quad (16)$$

Above, $\kappa_j$ is the $j$-th order cumulant, $\phi(x)$ is the pdf of a standard normal distribution, and $H_k$ are the Hermite polynomials. Note for arbitrary random variable $Y$, we can always define $X$ to be $\frac{Y-E[Y]}{\sqrt{Var[Y]}}$ so that Edgeworth expansion can be applied. See [4] for more details on the Edgeworth expansion.

The first four cumulants for the random variable $\sum_{k=1}^{u}\lambda_k T_k$ in (14) can be computed rather easily. Since $T_k \sim \chi^2_{r_k}$, $k=1,\dots,u$ we have

$$\kappa_1 = \mathrm{E}\left[\sum_{k=1}^{u}\lambda_k T_k\right] = \sum_{k=1}^{u}\lambda_k r_k$$

$$\kappa_2 = \mathrm{Var}\left[\sum_{k=1}^{u}\lambda_k T_k\right] = \sum_{k=1}^{u}\lambda_k^2 \mathrm{Var}\left[T_k\right] = \sum_{k=1}^{u} 2\lambda_k^2 r_k$$

$$\kappa_3 = \sum_{k=1}^{u} 8\lambda_k^3 r_k \qquad \kappa_4 = \sum_{k=1}^{u} 48\lambda_k^4 r_k.$$

The use of the approximation (15) leads to the following inversion of the probabilistic constraint $P(a^\top\theta \le b) \ge \eta$

$$B = \left\{\alpha : \Phi\left(\frac{-\kappa_1}{\sqrt{\kappa_2}}\right) - \frac{\kappa_3}{6}H_2\left(\frac{-\kappa_1}{\sqrt{\kappa_2}}\right)\phi\left(\frac{-\kappa_1}{\sqrt{\kappa_2}}\right) \ge \eta\right\}$$

where $\Phi$ is the cumulative density function (cdf) of a standard normal distribution. The derivation follows from the fact that $\phi^{(n)}(x) = (-1)^n H_n(x)\phi(x)$.

In theory, function (14) can be approximated to arbitrary precision by using higher order cumulants in the Edgeworth expansion. For random variable $\sum_{k=1}^{u}\lambda_k T_k$ in (14), its higher order cumulants have simple forms which again depend on $r_1,\dots,r_u$. This implies that the set $B$ can be approximated arbitrarily closely at very little computational cost. In practice, approximations such as (15) and (16) that use only the first four cumulants are often considered adequate and usually work well.

## 4 Gaussian Prior

The most popular prior for continuous unbounded parameters $\theta \in \mathbb{R}^n$ is the Gaussian distribution. It is often used in conjunction with a Gaussian model $f(z|\theta) = N(\theta, \Upsilon)$, $g(\theta|\mu,\Sigma) = N(\mu,\Sigma)$ where the posterior $p(\theta|D)$ is Gaussian as well. In this case the posterior and various integrals over it have a close form.

In the conditional or discriminative setting, a Gaussian prior is often used in conjunction with linear regression

$$f(y|x,\theta) = N(\theta^\top x, \sigma^2) \quad y \in \mathbb{R} \quad (17)$$

or logistic regression

$$f(y|x,\theta) = \left(1 + e^{-y\theta^\top x}\right)^{-1} \quad y \in \{-1,+1\}. \quad (18)$$

In both cases (17)-(18) a Gaussian prior over $\theta$ is the most popular means of introducing domain knowledge or regularizing the model.

Specifying domain knowledge by constraining $\theta$ is relatively easy as $\theta_1,\dots,\theta_n$ correspond directly to the expected values of the data dimensions $z_1,\dots,z_n$. For example, consider modeling a physical population quantity using a mixture of Gaussians. There may be reasons to believe that some mixture components correspond to specific groups in the population, enabling the use of domain knowledge to constrain the parameters of the mixture components. If the constraints are uncertain, introducing probabilistic rather than hard constraints will be more robust in the event of their failure.

In the conditional case, constraints on $\theta$ may reflect the relationship among the data and the predictor variable $y$. For example in a logistic regression model for classifying document topics, we may enforce $|\theta_i| \le c$ for some $i$ corresponding to stop-words or non-content words. The assumption that non-content words such as `the` or `of` do not contribute to the topic is a reasonable one. However, there are cases in which the constraints may not hold which motivate $\eta < 1$.

We turn now to inverting the constraints (2) and identifying the sets $B_1,\dots,B_n$ if $\theta \sim N(\mu,\Sigma)$. We have $u \stackrel{\text{def}}{=} a_i^\top \theta \sim N(\bar{u}, \sigma^2)$ where $\bar{u} = a_i^\top \mu$, $\sigma^2 = a_i^\top \Sigma a_i$ and

$$\begin{aligned} P\left(a_i^\top \theta \le b_i\right) \ge \eta_i &\Leftrightarrow P\left(\frac{u-\bar{u}}{\sigma} \le \frac{b_i - \bar{u}}{\sigma}\right) \ge \eta_i \\ &\Leftrightarrow \frac{b_i - \bar{u}}{\sigma} \ge \Phi^{-1}(\eta_i) \qquad (19) \\ &\Leftrightarrow a_i^\top \mu + \Phi^{-1}(\eta_i)\sqrt{a_i^\top \Sigma a_i} \le b_i \end{aligned}$$

($\Phi$ is the standard normal cdf). Further details concerning this derivation may be found in [2].

Depending on the problem structure, we may assume the hyperparameter $\alpha$ to be $(\mu, \Sigma)$ or just $\mu$ ($\Sigma$ is considered fixed in this case). One difficulty is that the MAP or EB optimization problem is specified in terms of $\Sigma^{-1}$ while the inverted constraints (19) are specified in terms of $\Sigma$. This difficulty is not substantial if $\Sigma$ is diagonal as $\Sigma^{-1} = \text{diag}(1/\Sigma_{11}, \ldots, 1/\Sigma_{nn})$.

In situations when $\Sigma$ is not a diagonal matrix obtaining the EB or MAP estimator subject to the inverted constraints is highly non-trivial from an optimization perspective. We propose instead to optimize a surrogate objective function based on the method of Bregman divergences. More details appear in Appendix A.

## 5 Experiments

We demonstrate our framework using experiments on synthetic and real-world data. The synthetic data experiments test the applicability of the framework to the multinomial, Gaussian, and linear regression cases. The real world experiments test the applicability of the framework to two NLP tasks: sentiment prediction and readability prediction, both using linear regression.

### 5.1 Synthetic Data Experiments

We start by evaluating the framework on the problem of estimating multinomial parameters under ordering constraints. We sampled data from $\text{Mult}(\theta)$ for $\theta = \left(\frac{1}{12}, \frac{1}{6}, \frac{1}{6}, \frac{1}{4}, \frac{1}{3}\right)$ and enforced the probabilistic constraints $A = \{\theta_i \le \theta_j,\ i = 1,2,3 \text{ and } j = 4,5\}$ (Figure 2, top left) and $B = \{\theta_i \ge \theta_j,\ i = 1,2,3 \text{ and } j = 4,5\}$ (Figure 2, bottom left). We used in this and other experiments (unless noted otherwise) a confidence value of $\eta_i = 0.95$. We assumed a Dirichlet prior for $\theta \sim \text{Dir}(\alpha)$, and a uniform hyperprior for $\alpha$.

In the Gaussian case we generated data from three normal distributions $N(\theta_1, 1), N(\theta_2, 1), N(\theta_3, 1)$ for $\theta = (\theta_1, \theta_2, \theta_3) = (0, 1/2, 1)$ and enforced the probabilistic constraints $C = \{\theta_1 \le \theta_2, \theta_2 \le \theta_3, \theta_1 \ge 0, \theta_3 \le 1\}$ (Figure 2, top middle) and $D = \{\theta_1 \ge \theta_2, \theta_2 \ge \theta_3\}$ (Figure 2, bottom middle).

In the case of linear regression, the samples were drawn from the model $y \sim N(\beta^\top x, 1)$ where $\beta$ is a 10 dimensional randomly generated vector whose first and last 5 components are uniformly distributed on $(-1, 0)$ and $(0, 1)$ respectively. We enforced the probabilistic constraints $E = \{\{\beta_1, \beta_3, \beta_5\} \le 0, \{\beta_6, \beta_8, \beta_{10}\} \ge 0, \{\beta_2, \beta_4\} \le \{\beta_7, \beta_9\}\}$ (Figure 2, top right) and $F = \{\{\beta_1, \beta_3, \beta_5\} \ge 0, \{\beta_6, \beta_8, \beta_{10}\} \le 0, \{\beta_2, \beta_4\} \ge \{\beta_7, \beta_9\}\}$ (Figure 2, bottom right). In this case we applied ridge regularization to both the MLE and the constrained MLE and report the best results from the following set of ridge parameters $\{0.001, 0.01, 0.1, 0.2, 0.5, 1, 2, 5, 10, 100\}$.

In all three cases we observe similar results. When the constraints are correct (top row) incorporating them via constrained MLE (hard constraints) or MAP, EB (probabilistic constraints) provides higher estimation accuracy over the non-constrained MLE.

However, when some of the constraints are inaccurate, incorporating them as hard constraints hurts performance substantially and results in much poorer estimation as compared to the unconstrained MLE. This is to be expected as hard inaccurate constraints force the estimator away from the true parameters. On the other hand, incorporating inaccurate probabilistic constraints using MAP or EB performs remarkably well with almost equal performance to the unconstrained MLE. The inaccurate constraints don't hurt the estimator as the constraints are simply ignored due to their clash with the information embedded in the data. Note that this holds even for high confidence values such as $\eta = 0.95$ (our choice for these experiments).

### 5.2 Sentiment and Readability Prediction

To test the validity of the framework on real world data we experimented with two NLP tasks: sentiment and readability prediction where the underlying model is linear regression.

For sentiment prediction, we randomly chose 2 out of 4 movie critics from the Cornell sentiment scale datasets[1], which results in collections of 1027 and 1307 documents respectively, with 4 sentiment levels ranging from 1 (very bad) to 4 (very good). For readability prediction, we used the weekly reader dataset, obtained by crawling the Weekly Reader[2] commercial website after receiving special permission. The readability dataset contains a total of 1780 documents, with 4 readability levels ranging from 2 to 5 indicating the school grade levels of the intended audience. Preprocessing includes lower-casing, stop word removal, stemming, and selecting 1000 top features based on document frequency. The predictor variable is also centered for ease of applying parameter constraints.

[1] http://www.cs.cornell.edu/People/pabo/movie-review-data

[2] www.wrtoolkit.com

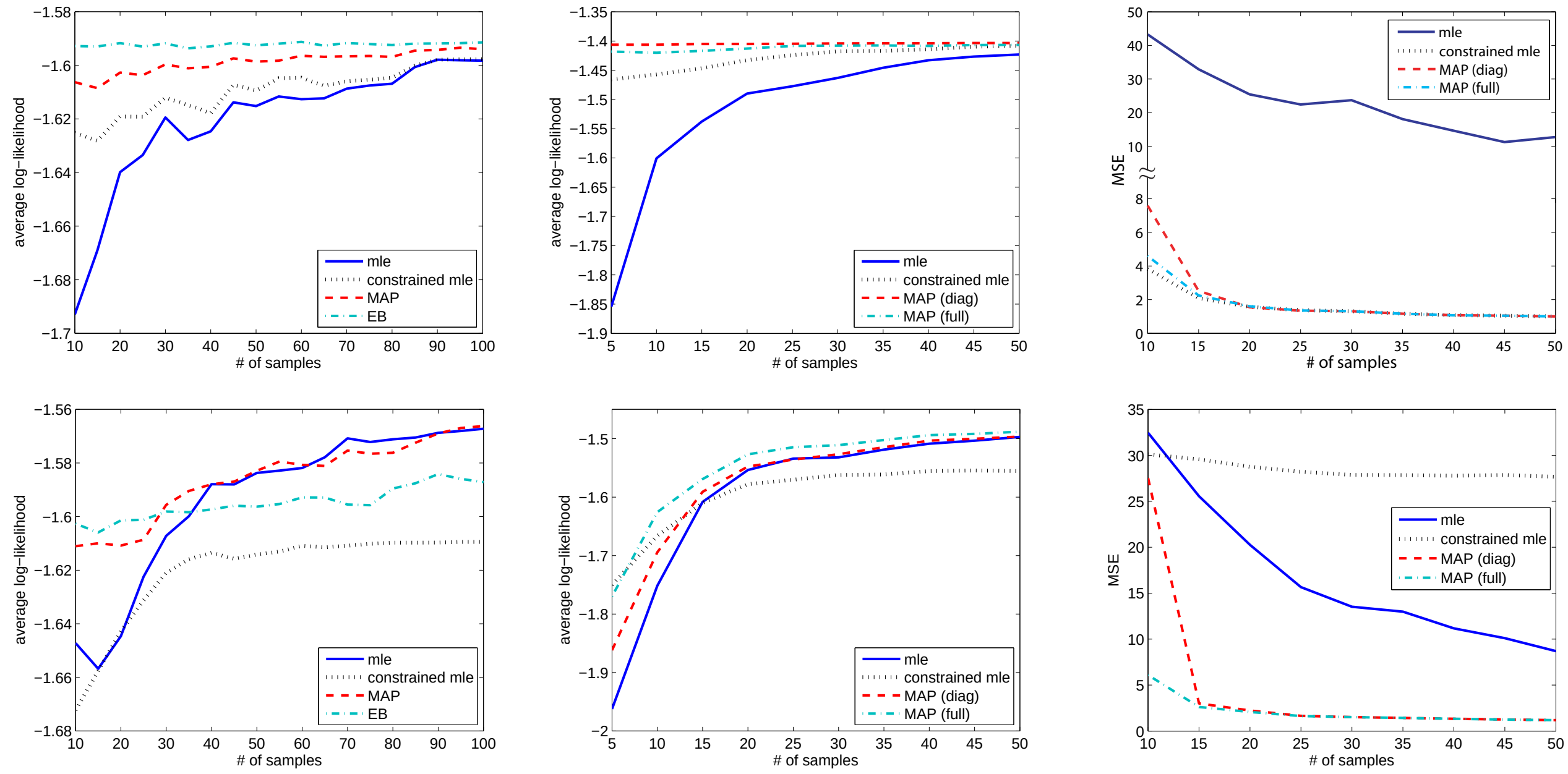

Figure 2: Average test set performance for multinomial (left), Gaussian (middle) and linear regression (right) over 20 random train/test splits. Multinomial parameters are estimated by MLE, (hard) constrained MLE, and probabilistic constraint EB and MAP. Gaussian means and regression parameters are estimated by MLE, (hard) constrained MLE and probabilistic constraint MAP with either diagonal or full covariance matrix. Ridge regularization is applied to both MLE and constrained MLE for linear regression. In all three cases, the top row corresponds to correct constraints while the bottom row corresponds to incorrect constraints.

The probabilistic constraints for the sentiment prediction experiment were developed by one of the authors after being presented with the vocabulary of the dataset. The author was asked to pick two subsets of the vocabulary - one associated with positive sentiment and one with negative sentiment. A total of 190 and 154 words were chosen for the two critics. A subset of these words starting with 'a' are listed in Table 1. For words that are deemed indicative of positive sentiment, we enforce $\theta_i \geq b$ for some nonnegative number $b$ as the parameter constraints. Similarly, we enforce $\theta_i \leq -b$ for the negative words.

In the case of readability prediction, we assume that the appearance of longer words implies higher readability level than the appearance of shorter words. To this end, we randomly chose 600 pairs of words of different length, and required that the parameters corresponding to the longer words have a higher value than the parameters corresponding to the shorter words. Note that in both the sentiment and readability cases the constraints represent reasonable domain knowledge but may not be entirely accurate.

Figure 3 compares ridge regression, constrained ridge regression and MAP with a full covariance matrix. We chose to use a full rather than a diagonal matrix due to the correlation between the regression parameters. The ridge parameter was chosen from the set $\{0.2, 0.5, 1, 2, 5\}$. Variance $\sigma^2$ of the linear regression model (17) is assumed to lie in $\{0.5, 1, 2\}$ and $\Lambda$ in (20) takes the form of $\tau I$ where $\tau$ is chosen from $\{0.01, 0.05, 0.1, 0.2, 0.3\}$. The parameter value bound $b$ in sentiment prediction is set to be 0.

| | | | |
|---|---|---|---|
| *appeal*[1,2] | *award*[1,2] | *accomplish*[1,2] | *attract*[1,2] |
| *amus*[1,2] | annoi[1,2] | *appar*[1] | avoid[1] |
| *adequ*[1] | *amaz*[1] | aw[1] | *appreci*[1] |
| awkward[2] | absurd[2] | *achiev*[2] | artifici[2] |
| *art*[2] | arti[2] | *admir*[2] | |

Table 1: Words chosen for parameter constraints for sentiment prediction. Superscript numbers indicate the movie critic. Italics blue words indicate positive sentiment while non-italics black words indicate negative sentiment.

The results shown in Figure 3 illustrate that the probabilistic constraints help improve accuracy over the unconstrained MLE. More impressive is the fact that they result in a substantial improvement in modeling accuracy also over the (hard) constrained MLE. This is due to the uncertain nature of the constraints and the fact that some of the constraints do not hold. This underscores the main point of the paper that domain knowledge is often uncertain and better enforced using probabilistic parameter constraints rather than hard ones.

It is worth mentioning that the framework is not sensitive to the choice of $\eta$ for a broad region of possible $\eta$. For sentiment prediction, we have experimented with $\eta$ being equal to 0.75, 0.85 and 0.95, and parameter value bound $b$ being equal to 0, 0.1 and 0.5. For all those combinations of parameters, we found that the graphs are quite similar to Figure 3 except for individual values of likelihood or accuracy. This is indeed a desirable property for real-world applications when the confident level expressed by a domain expert may be subject to uncertainty.

## 6 Discussion

Incorporating knowledge into the learning process has been studied extensively by the statistics community. Frequentists use it to define the model and constrain the parameter space. Bayesians use it to define the model and the prior over the parameter space. In the Bayesian case, uncertainty is usually handled by using hierarchical models with diffuse hyperpriors [1]. Obtaining domain knowledge is addressed by prior elicitation in the subjective Bayes community [7].

Our work differs from the standard prior elicitation approach in that we do not elicit the prior directly. Rather we elicit parameter constraints and confidence values which are used in turn to derive an equivalent prior in a hierarchical Bayes setting via Proposition 1. Standard Bayesian inference can then proceed on the equivalent model in the usual manner.

The advantage of doing so is that it is much easier for domain experts to specify constraints and confidence values. Directly specifying a prior is considerably less intuitive as it makes it hard to discern the confidence with which specific assertions are made. Thus, our contribution is in nicely separating the domain assertions and their confidence values in a simple and intuitive way.

In the machine learning community, parameter uncertainty has been addressed by a variety of techniques, many of which are algorithmic in nature. Related research on incorporating parameter uncertainty are [6, 5] which consider a linear classifier with a Gaussian prior over the model parameter, and update the hyperparameters online using probabilistic parameter constraints.

Using experiments on synthetic and real world data we show that uncertain domain knowledge can be effectively incorporated in practice. The use of uncertain constraints leads to high modeling accuracy when the constraints are accurate. In case the constraints are inaccurate, the uncertainty prevents the model from performing poorly which stands in contrast to hard constraints that push the parameters away from their true values.

Specifying domain knowledge in real world situations is sometimes bound to be inaccurate. Our approach enables the use of a large number of domain statements without worrying too much about the validity of each specific statement. Our experiments indicate that it works well for natural language problems where domain knowledge is relatively easy to specify. It is likely that the framework performs similarly well in other areas where domain knowledge is available and the underlying model has a Dirichlet or Gaussian prior.

### Acknowledgements

This research was supported in part by NSF grants IIS-0906550 and DMS-0604486.

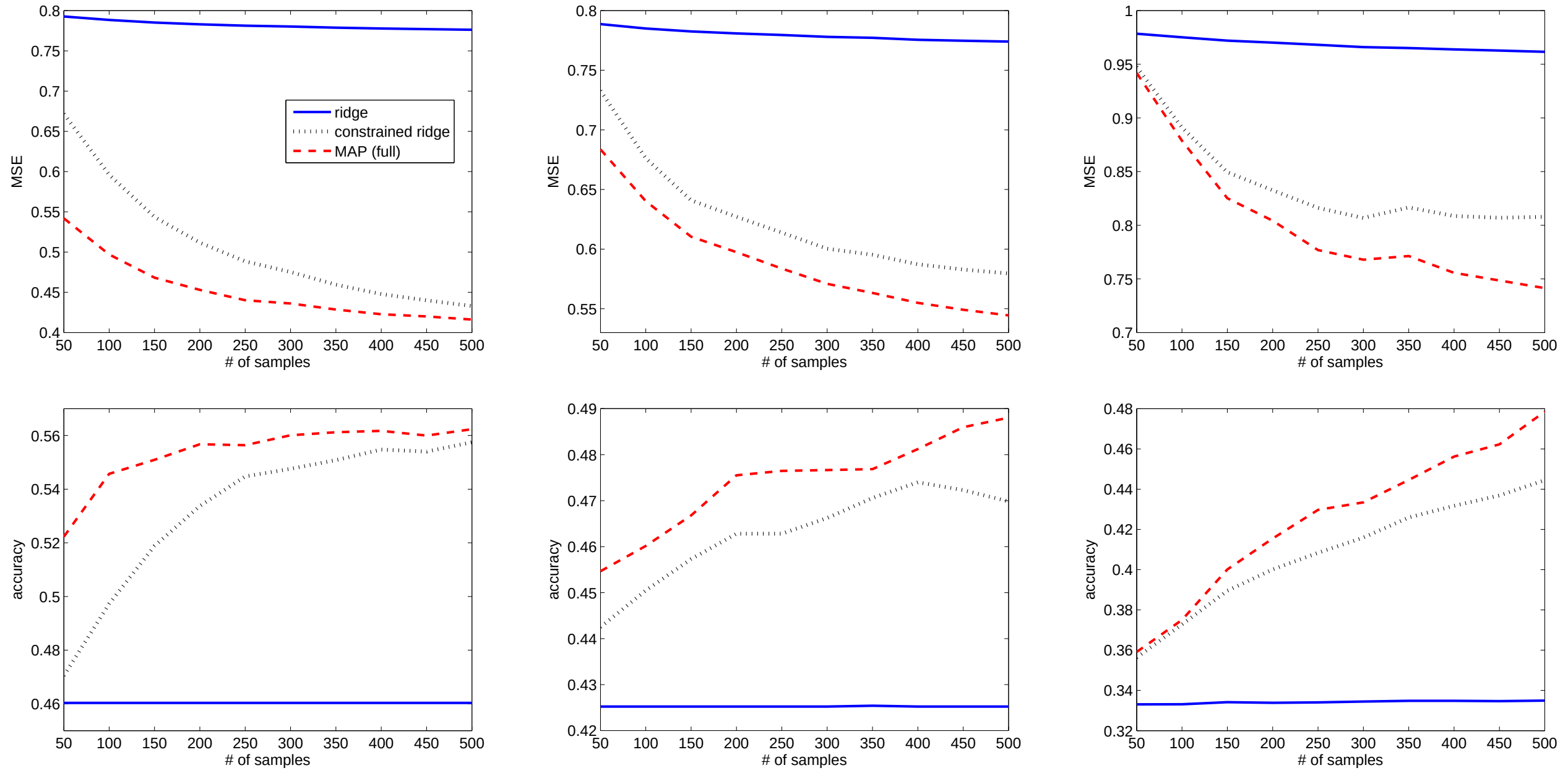

Figure 3: Test-set mean square error (MSE, top) and accuracy rates (bottom) over 10 iterations for sentiment prediction (left, middle corresponding to the two critics) and readability prediction (right). Regression parameters are estimated by ridge, (hard) constrained ridge and probabilistic constraint MAP with full covariance matrix.

# A Bregman Projection

We consider the problem of obtaining the MAP for $(\theta, \alpha)$, with $\alpha = (\mu, \Sigma)$, $\theta \sim N(\mu, \Sigma)$ in the case of linear constraints with a non-diagonal $\Sigma$. We make a standard assumption regarding the hyperprior $h(\alpha)$

$$h\left(\mu, \Sigma^{-1}\right) \propto \exp\left(-\frac{1}{2}\mathrm{tr}\left(\Sigma^{-1}\Lambda\right)\right) \qquad (20)$$

($\Lambda$ is a positive definite matrix) which is equivalent to stating that $\Sigma^{-1} \sim \mathrm{Wishart}_{n+1}\left(\Lambda^{-1}\right)$, $\mu|\Sigma^{-1}$ is uniform. Note that the techniques introduced below apply to arbitrary $f(\cdot|\theta)$ as no assumptions are made regarding the particular choice of $f$.

In the case of the non-diagonal $\Sigma$ solving the constrained MAP problem (10) is highly non-trivial. We propose to use an iterative optimization technique, and during the step of optimizing $\Sigma$ with fixed $\theta$ and $\mu$, maximize instead a surrogate objective function based on the method of Bregman projection. Specifically, we solve the following problem to obtain the point estimator for $\Sigma$ for fixed $\theta$ and $\mu$

$$\min_{\Sigma} \quad D_{\mathrm{LogDet}}\left(\Sigma, \Lambda + (\theta-\mu)(\theta-\mu)^\top\right) \qquad (21)$$

$$\text{s.t.} \quad \mathrm{tr}\left(\Sigma a_i a_i^\top\right) \le \left(\frac{b_i - a_i^\top \mu}{\Phi^{-1}(\eta_i)}\right)^2, \quad i = 1, \ldots, l.$$

The divergence above is the LogDet Bregman divergence between matrices [3]. The hyperparameter $\Sigma$ estimated by (21) is then used when we subsequently optimize over $\theta$ or $\mu$.

The problem (21) has the same constraints as the original subproblem but a different objective function which is originally $D_{\mathrm{LogDet}}(\Lambda + (\theta-\mu)(\theta-\mu)^\top, \Sigma)$. By switching the arguments of the objective function, we are able to solve the problem using the method of Bregman projections [3] which achieves the optimal solution by sequentially projecting to different convex regions defined by the corresponding constraints.

A useful property of Bregman projection is that it can be used to ensure the positive definiteness of $\Sigma^{-1}$ and $\Sigma$ when starting from a positive definite matrix $\Lambda$. This results immediately from the fact that each update of $\Sigma^{-1}$ by projecting the matrix divergence onto the convex region defined by $\mathrm{tr}\left(\Sigma a_i a_i^\top\right) \le z_i$ takes the following form $\Sigma^{-1} = \left(\Lambda + (\theta-\mu)(\theta-\mu)^\top\right)^{-1} + \nu a_i a_i^\top$ where $\nu = \max\left\{0, \frac{a_i^\top(\Lambda+(\theta-\mu)(\theta-\mu)^\top)a_i - z_i}{z_i a_i^\top(\Lambda+(\theta-\mu)(\theta-\mu)^\top)a_i)}\right\} \ge 0$. Detailed derivations are omitted due to lack of space.